\documentclass[10pt,twocolumn,letterpaper]{article}

\def \arxiv {} 
\ifx \arxiv \undefined
    \usepackage{cvpr}              
\else
    \usepackage[pagenumbers]{cvpr} 
\fi

\usepackage{epsfig}
\usepackage{graphicx}
\usepackage{amsmath}
\usepackage{amssymb}
\usepackage{booktabs}
\usepackage{lib}
\usepackage{mathtools}

\usepackage{xspace}
\usepackage[accsupp]{axessibility}  

\makeatletter
\DeclareRobustCommand\onedot{\futurelet\@let@token\@onedot}
\def\@onedot{\ifx\@let@token.\else.\null\fi\xspace}

\def\eg{\emph{e.g}\onedot} 
\def\ie{\emph{i.e}\onedot} 
 
\def\etc{\emph{etc}\onedot} 
 
\def\etal{\emph{et al}\onedot}
\makeatother
\usepackage{stmaryrd} 
\usepackage{paralist} 
\usepackage[symbol]{footmisc}

\usepackage[pagebackref,breaklinks,colorlinks,urlcolor=red]{hyperref}

\usepackage{xfrac}  

\usepackage[capitalize]{cleveref}
\crefname{section}{Sec.}{Secs.}
\Crefname{section}{Section}{Sections}
\Crefname{table}{Table}{Tables}
\crefname{table}{Tab.}{Tabs.}

\usepackage{color}
\definecolor{crimson}{rgb}{0.86, 0.08, 0.24}
\definecolor{gray}{rgb}{0.5,0.5,0.5}
\definecolor{green}{rgb}{0, 0.4, 0}
\definecolor{orange}{rgb}{1, 0.5, 0}
\definecolor{mahogany}{rgb}{0.75, 0.25, 0.0}
\definecolor{purple}{rgb}{0.6, 0, 0.6}
\definecolor{darkgreen}{rgb}{0, 0.4, 0}
\definecolor{frenchblue}{rgb}{0.0, 0.45, 0.73}
\definecolor{blue}{rgb}{0.0, 0.0, 0.65}
\definecolor{red}{rgb}{1,0,0}
\definecolor{yellow}{rgb}{1,1,0}
\definecolor{magenta}{rgb}{1,0,1}
\definecolor{pink}{rgb}{1,0.412,0.706}
\definecolor{newgreen}{rgb}{0, 0.6, 0.2}

\newlength\paramargin
\newlength\figmargin
\newlength\subfigmargin
\newlength\subsecmargin
\newlength\tabmargin
\newlength\eqmargin

\newlength\presecmargin
\newlength\secmargin

\newlength\rulelength
\newcommand{\comment}[1]{}

\newcommand{\subsecref}[1]{Section~\ref{subsec:#1}}
\newcommand{\tabref}[1]{Table~\ref{tab:#1}}

\ifx \submission \undefined

\else

\fi

\def\cvprPaperID{367} 
\def\confName{CVPR}
\def\confYear{2023}

\newcommand{\modelName}{SDFusion\xspace}

\begin{document}

\title{\modelName: Multimodal 3D Shape Completion, Reconstruction, and Generation}


\author{
Yen-Chi Cheng$^{1}$
\hspace*{1em}
Hsin-Ying Lee$^{2}$
\hspace*{1em}
Sergey Tulyakov$^{2}$
\hspace*{1em}
Alexander Schwing$^{1*}$
\hspace*{1em}
Liangyan Gui$^{1*}$
\\
$^{1}$University of Illinois Urbana-Champaign
\hspace*{1em}
$^{2}$Snap Research
\\
\small \{\texttt{yenchic3,aschwing,lgui}\}\texttt{@illinois.edu}
\hspace*{1em}
\{\texttt{hlee5,stulyakov}\}\texttt{@snap.com}
\\
\small \url{https://yccyenchicheng.github.io/SDFusion/}
}

\twocolumn[{%
\maketitle
\vspace{-1.0em}
\renewcommand\twocolumn[1][]{#1}%
    \centering 
    \vspace{-3.5mm}
    \includegraphics[width=.98\linewidth]{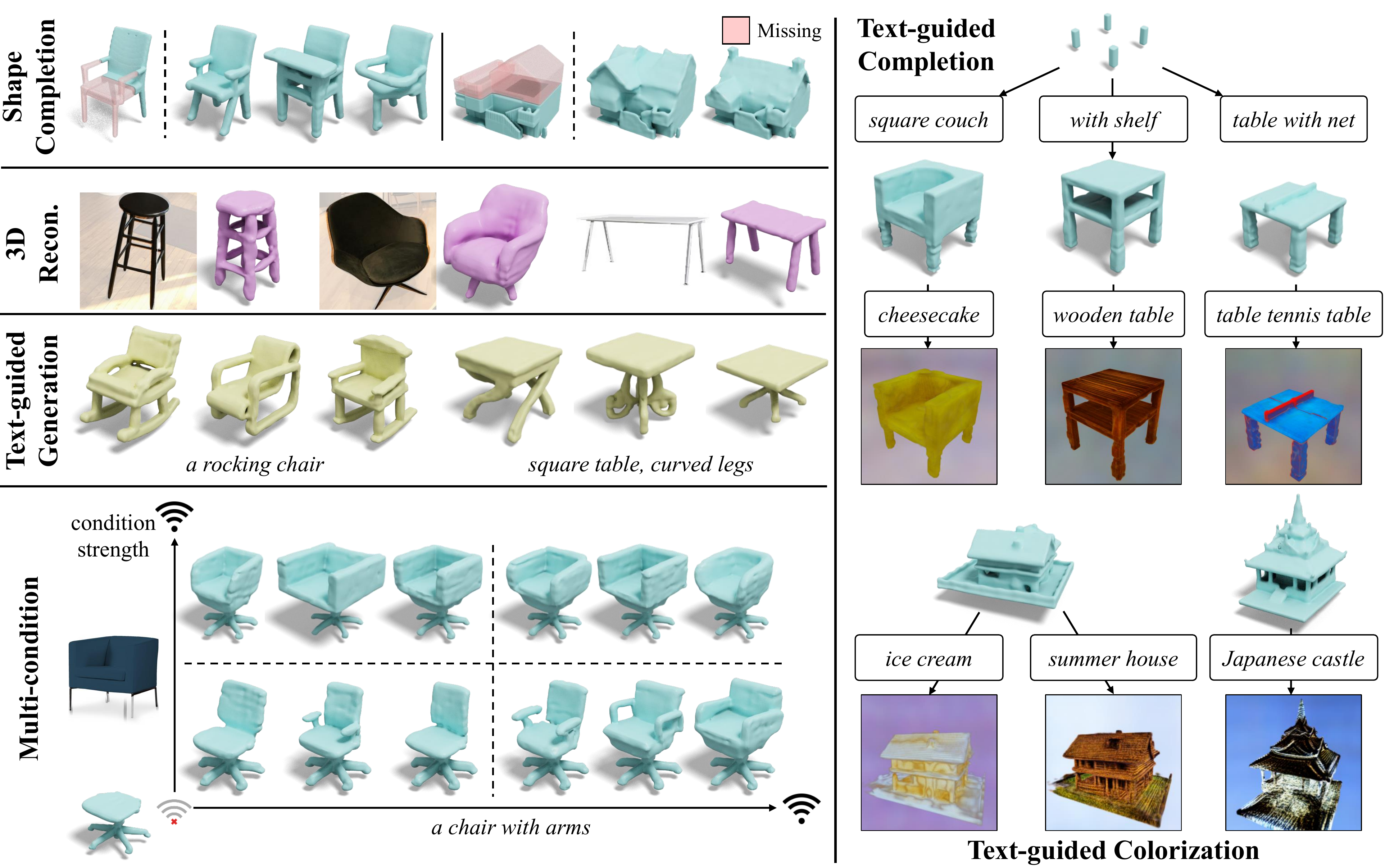}
    \vspace{-0.2cm}
    \captionof{figure}{
        \textbf{Applications of \modelName.}
        The proposed diffusion-based model enables various applications.
        (\textit{left}) \modelName can generate shapes conditioned on different input modalities, including partial shapes, images, and text. \modelName can even jointly handle multiple conditioning modalities while controlling the  strength for each of them.
        (\textit{right}) We leverage pretrained 2D models to texture 3D shapes generated by \modelName.
    } \figlabel{teaser}
    \vspace{2mm}
}]


\begin{abstract}
\vspace{-.5cm}
    In this work, we present a novel framework built to simplify 3D asset generation for amateur users. To enable interactive generation, our method supports a variety of input modalities that can be easily provided by a human, including images, text, partially observed shapes and combinations of these, further allowing to adjust the strength of each input. At the core of our approach is an encoder-decoder, compressing 3D shapes into a compact latent representation, upon which a diffusion model is learned. To enable a variety of multi-modal inputs, we employ task-specific encoders with dropout followed by a cross-attention mechanism. Due to its flexibility, our model naturally supports a variety of tasks, outperforming prior works on shape completion, image-based 3D reconstruction, and text-to-3D. Most interestingly, our model can combine all these tasks into one swiss-army-knife tool, enabling the user to perform shape generation using incomplete shapes, images, and textual descriptions at the same time, providing the relative weights for each input and facilitating interactivity. Despite our approach being shape-only, we further show an efficient method to texture the generated shape using large-scale text-to-image models. 
\end{abstract}

\newcommand{\val}{\mathbf{v}}
\newcommand{\pos}{\mathbf{x}}
\newcommand{\posg}{g}
\newcommand{\R}{\mathbb{R}}


\section{Introduction}
\seclabel{intro}




Generating 3D assets is a cornerstone of immersive augmented/virtual reality experiences. Without realistic and  diverse objects, virtual worlds will look void and engagement will remain low. Despite this need, manually creating and editing 3D assets is a notoriously difficult task, requiring creativity, 3D design skills, and access to sophisticated software with a very steep learning curve. This makes 3D asset creation inaccessible for inexperienced users. Yet, in many cases, such as interior design, users more often than not have a reasonably good understanding of what they want to create. In those cases, an image or a rough sketch is sometimes accompanied by text indicating details of the asset, which are hard to express graphically for an amateur. 

\comment{Conventional 3D generative models learn 3D modeling using direct 3D supervision, like point clouds~\cite{achlioptas2018learning,luo2021diffusion}, SDF~\cite{chen2019learning,autosdf2022}, voxels~\cite{smith2017improved,xie2018learning}, \etc. 
Recently, first efforts have been made to explore  learning of 3D geometry from only 2D supervision by incorporating inductive biases via neural rendering techniques~\cite{chan2021pi,schwarz2020graf,chan2022efficient,gu2021stylenerf}.}

Due to this need, it is not surprising that democratizing the 3D content creation process has become an active research area. 
Conventional 3D generative models require direct 3D supervision in the form of point clouds~\cite{achlioptas2018learning,luo2021diffusion}, signed distance functions (SDFs)~\cite{chen2019learning,autosdf2022}, voxels~\cite{smith2017improved,xie2018learning}, \etc. 
Recently, first efforts have been made to explore the learning of 3D geometry from multi-view supervision with known camera poses by incorporating inductive biases via neural rendering techniques~\cite{chan2021pi,schwarz2020graf,chan2022efficient,gu2021stylenerf,ZhaoECCV2022a}.
While compelling results have been demonstrated, training is often very time-consuming and ignores available 3D data that can be used to obtain good shape priors.
We foresee an ideal collaborative paradigm for generative methods where models trained on 3D data provide detailed and accurate geometry, while models trained on 2D data provide diverse appearances.
A first proof of concept is shown in \figref{teaser}.
In our pursuit of flexible and high-quality 3D shape generation, we introduce \textbf{\modelName}, a diffusion-based generative model with a signed distance function (SDF) under the hood, acting as our 3D representation. 
Compared to other 3D representations, SDFs are known to represent well high-resolution shapes with arbitrary topology\cite{jiang2020sdfdiff,chen2019learning,mescheder2019occupancy,park2019deepsdf}.
However, 3D representations are infamous for demanding high computational resources, limiting most existing 3D generative models to voxel grids of $32^3$ resolution  and point clouds of $2K$ points.
To side-step this issue, we first utilize an auto-encoder to compress 3D shapes into a more compact low-dimensional representation.
Because of this, \modelName can easily scale up to a $128^3$ resolution.
To learn the probability distribution over the introduced latent space, we leverage diffusion models, which have recently been used with great success in various 2D generation tasks~\cite{meng2021sdedit,sinha2021d2c,avrahami2022blended,nichol2021glide,kawar2022imagic,saharia2022photorealistic}. 
Furthermore, we adopt task-specific encoders and a cross-attention~\cite{rombach2022high} mechanism to support multiple conditioning inputs, and apply  classifier-free guidance~\cite{ho2022classifier} to enable flexible conditioning usage. 
Because of these strategies, \modelName can not only use a variety of conditions from multiple modalities, but also  adjust their importance weight, as shown in \figref{teaser}.
Compared to a recently proposed autoregressive model~\cite{autosdf2022} that also adopts an encoded latent space, \modelName achieves superior sample quality, while offering more flexibility to handle multiple conditions and, at the same time, features reduced memory usage.
With \modelName, we study the interplay between models trained on 2D and 3D data.
Given 3D shapes generated by \modelName, we take advantage of an off-the-shelf 2D diffusion model~\cite{rombach2022high}, neural rendering~\cite{mildenhall2020nerf}, and score distillation sampling~\cite{poole2022dreamfusion} to texture the shapes given text descriptions as conditional variables. 

\comment{Diffusion models have shown great success in generative modeling. But adopting them in learning the 3D shapes directly is computationally intractable as they live in a complex and high-dimensional 3D space. We utilize auto-encoders to first compress 3D shapes into a low-dimensional and compact representations. This latent space is expressive enough and makes the training of diffusion models feasible. While previous work \todo{cite autosdf, shapeformer} propose to use autoregressive models, using diffusion models has the following advantage -- (1) memory efficient during training (time dependent UNet)? (2) higher quality. \todo{compare with autoregressive models}. To model the conditional distribution, we use task-specific encoder to encode the inputs such as images, texts, and use the cross-attention as the conditional mechanism. Finally, with the generated shape, we learn a neural radiance field to add textures onto the shape. We adopt differentiable rendering and an off-the-shelf 2D diffusion model to provide the guidance of how to update the neural radiance field.}

We conduct extensive experiments on the ShapeNet~\cite{shapenet2015}, BuildingNet~\cite{selvaraju2021buildingnet}, and Pix3D~\cite{sun2018pix3d} datasets. We show that \modelName quantitatively and qualitatively outperforms prior work in shape completion, 3D reconstruction from images, and text-to-shape tasks.
We further demonstrate the capability of jointly controlling the generative model via multiple conditioning modalities, the flexibility of adjusting relative weight among modalities, and the ability to texture 3D shapes given  textual descriptions, as shown in \figref{teaser}. 
 
We summarize the main contributions as follows:
\begin{compactitem}
    \item We propose \modelName, a diffusion-based 3D generative model which uses a signed distance function as its 3D representation and a latent space for diffusion.
    \item \modelName enables conditional generation with multiple modalities, and  provides flexible usage by adjusting the weight among modalities.
    \item We demonstrate a pipeline to synthesize textured 3D objects benefiting from an interplay between 2D and 3D generative models. 
\end{compactitem}

\begin{figure*}[t!]
    \centering
     \includegraphics[width=0.985\linewidth]{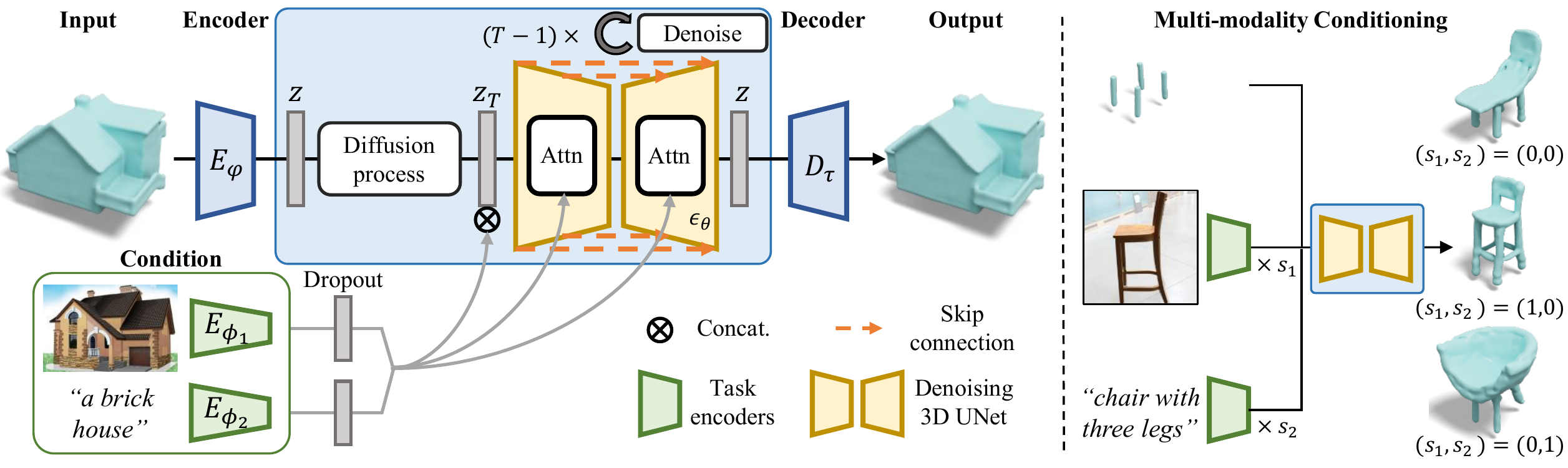}
     \vspace{-3mm}
    \caption{
    \textbf{\modelName Overview.}
    (left) To enable high-resolution generation, we first encode 3D shapes into a latent space, where a diffusion model is trained. Furthermore, to enable flexible conditional generation, we adopt class-specific encoders along with classifier-free guidance to enable multi-modality conditioning. 
    (right) At inference time, we can control the importance of each conditioning modality.
    }
    \vspace \figmargin
    \vspace{-3mm}
    \figlabel{overview_p1}
\end{figure*} 

\section{Related Work}
\vspace{-2mm}
\seclabel{related_work}
\vspace{-0.5mm}
\paragraph{3D Generative Models.}
Different from 2D images, it is less clear how to effectively represent 3D data. Indeed,  various representations with different pros and cons have been explored, particularly when considering 3D generative models. 
For instance, 3D generative models have been explored for  point clouds~\cite{achlioptas2018learning,luo2021diffusion}, voxel grids~\cite{smith2017improved,xie2018learning,lin2023infinicity}, meshes~\cite{zhang2021sketch2model}, signed distance functions (SDFs)~\cite{chen2019learning,autosdf2022,cheng2022cross}, \etc.
In this work, we aim to generate an SDF. Compared to other representations, SDFs exhibit a reasonable trade-off regarding expressivity, memory efficiency, and direct applicability to downstream tasks. Moreover, conditioning 3D generation of SDFs on different modalities further enables many applications, including shape completion, 3D reconstruction from images, 3D generation from text, \etc. The proposed framework can handle these tasks in a single model which makes it different from prior work.

Recently, thanks to the advancement of neural rendering~\cite{mildenhall2020nerf}, a new stream of research has emerged to learn 3D generation and manipulation from only 2D supervision~\cite{chan2021pi,schwarz2020graf,chan2022efficient,niemeyer2021giraffe,skorokhodov3d,xu2022discoscene,abdal20233davatargan,siarohin2023unsupervised}. We believe the interplay between two streams of work is promising in the foreseeable future.

\paragraph{Diffusion Models.}
Diffusion models have recently emerged as a popular family of generative models with competitive sample quality.
In particular, diffusion models have shown impressive quality, diversity, and expressiveness in various tasks such as image synthesis~\cite{ho2020denoising,ho2021cascaded,nichol2021improved,dhariwal2021diffusion}, super-resolution~\cite{saharia2022image}, image editing~\cite{meng2021sdedit,sinha2021d2c,cheng2022adaptively}, text-to-image synthesis~\cite{nichol2021glide,avrahami2022blended,kawar2022imagic,saharia2022photorealistic,rahman2022make}, \etc. 
In contrast to the flourishing research on diffusion models for 2D data, diffusion models have not yet been fully explored for 3D data. Notable exceptions include  attempts to apply diffusion models to point clouds~\cite{zhou20213d,luo2021diffusion}.

Differently, in this work, we apply diffusion models on SDF representations. As reasonable resolutions of SDFs are demanding to model, we study the use of a latent diffusion technique~\cite{rombach2022high} and the classifier-free conditional generation mechanism~\cite{ho2022classifier}, both of which have been shown to yield promising results when being used in 2D diffusion models.

\section{Approach}
\seclabel{approach}



We aim at synthesizing 3D shapes using diffusion models.
Towards this goal, we model the distribution over 3D shapes $\mathbf{X}$, a volumetric Truncated Signed Distance Field (T-SDF).
However, applying diffusion models directly on reasonably high-resolution 3D shapes is computationally very demanding. 
Therefore, we first compress the 3D shape into a discretized and compact latent space (\subsecref{vq}). This allows us to apply diffusion models in a lower-dimensional space (\subsecref{ldm}). 
The proposed framework can further incorporate various user conditions such as partial shapes, images, and text (\subsecref{poe}).
Finally, we showcase an interplay between the proposed framework and diffusion models trained on 2D data to texture 3D shapes (\subsecref{texture}).

\subsection{3D Shape Compression of SDF}
\label{subsec:vq}
A 3D shape representation is high-dimensional and thus difficult to model. To make  learning of high-resolution 3D shape distributions via diffusion models feasible, we compress the 3D shape $\bfX$ into a lower-dimensional yet compact latent space. For this, we leverage a 3D-variant of the Vector Quantised-Variational AutoEncoder (VQ-VAE)~\cite{oord2017neural}. 
Specifically, the employed 3D VQ-VAE contains an encoder $E_{\varphi}$ to encode the 3D shape into the latent space, and a decoder $D_{\tau}$ to decode the latent vectors back to 3D space. Given an input shape represented via the T-SDF $\bfX \in \mathbb{R}^{D \times D \times D}$, we have
\begin{equation}
    \begin{aligned}
        \bfz = E_{\varphi}(\bfX), \quad\text{and}\quad \bfX^{\prime} = D_{\tau}(\text{VQ}(\bfz)), 
    \end{aligned}
\end{equation}
where $\bfz \in \mathbb{R}^{d \times d \times d}$ is the latent vector, latent dimension $d$ is smaller than 3D shape dimension $D$, and $\text{VQ}$ is the quantization step which maps the latent variable $\bfz$ to the nearest element in the codebook $\mathcal{Z}$. The encoder $E_{\varphi}$, decoder $D_{\tau}$, and codebook $\mathcal{Z}$ are jointly optimized. 
We pre-train the VQ-VAE with reconstruction loss, commitment loss, and VQ objective, similar to~\cite{oord2017neural} using ShapeNet or BuildingNet data. 
\ifx \arxiv \undefined {
For more details please see the supplementary material.
} \fi

\subsection{Latent Diffusion Model for SDF}
\label{subsec:ldm}


Using the trained encoder $E_{\varphi}$, we can now encode any given SDF into a compact and low-dimensional latent variable $\bfz = E_{\varphi}(\bfX)$.
We can then train a diffusion model on this latent representation.
Fundamentally, a diffusion model learns to sample from a target distribution by reversing a progressive noise diffusion process.
Given a sample $\bfz$, we obtain $\bfz_{t}, t \in \{ 1, \ldots, T\}$ by gradually adding Gaussian noise with a variance schedule.
Then we use a time-conditional 3D UNet $\epsilon_\theta$ as our denoising model.
To train the denoising 3D UNet, we adopt the simplified objective proposed by Ho~\etal~\cite{ho2020denoising}:
\begin{equation}
    \begin{aligned}
        L_{\mathrm{simple}}(\theta) \coloneqq \mathbb{E}_{\bfz, \epsilon \sim N(0, 1), t} \left[ \norm{\epsilon - \epsilon_\theta(\bfz_t, t) }^2 \right].
    \end{aligned}
\end{equation}
At inference time, we sample $\widehat{\bfz}$ by gradually denoising a noise variable sampled from the standard normal distribution $N(0, 1)$, and leverage the trained decoder $D_{\tau}$ to map the denoised code $\widehat{\bfz}$ back to a 3D T-SDF shape representation $\widehat{\bfX} = D_{\tau} (\widehat\bfz)$, as shown in \figref{overview_p1}.

\subsection{Learning the Conditional Distribution}
\label{subsec:poe}

Being able to randomly sample shapes provides limited ability for interaction. Therefore, learning of a conditional distribution is essential for user applications. Importantly, multiple forms of conditional inputs are desirable such that the model can account for various kinds of  scenarios. Thanks to the flexible conditional mechanism provided by a latent diffusion model~\cite{rombach2022high}, we can incorporate multiple conditional input modalities at once with task-specific encoders $E_{\phi}$ and a cross-attention module. To further allow for more flexibility in controlling the distribution, we adopt classifier-free guidance for conditional generation. The objective function reads as follows: 
\begin{equation}
    \begin{aligned}
        L(\theta, \{\phi_i\}) \coloneqq \mathop\mathbb{E}_{\bfz, \bfc, \epsilon, t} \left[ \norm{\epsilon - \epsilon_\theta(\bfz_t, t, F\{D\circ E_{\phi_i}(\bfc_i)\}) }^2 \right],
    \end{aligned}
\end{equation}
where $E_{\phi_i}(\bfc_i)$ is the task-specific encoder for the $i^\text{th}$ modality, $D$ is a dropout operation enabling classifier-free guidance, and $F$ is a feature aggregation function. In this work, $F$ refers to a simple concatenation.

At inference time, given conditions from multiple modalities, we perform classifier-free guidance as follows
\begin{equation}
    \begin{aligned}
    &\epsilon_\theta(\bfz_t, t, F\{E_{\phi_i}~\forall i\}) = \epsilon_\theta(\bfz_t, t, \boldsymbol{\emptyset})+\\
    & \sum_i s_i\left(\epsilon_\theta(\bfz_t, t, F\{E_{\phi_i}(\bfc_i), E_{\phi_j}(\bfc_j) : \bfc_j= \boldsymbol{\emptyset} ~\forall j\neq i\})  \right.\\
    &\left.- \epsilon_\theta(\bfz_t, t, \boldsymbol{\emptyset})\right),
    \end{aligned}
\end{equation}
where $s_i$ denotes the weight of conditions from the $i^\text{th}$ modality and $\boldsymbol{\emptyset}$ denotes a condition filled with zeros. 
Intuitively, modalities with larger weights play more important roles in guiding the conditional generation. 

In this work, we study \modelName combined with three conditional modalities applied separately or jointly.
For \textit{shape completion}, given a partial observation of a shape, we  perform blended diffusion similar to~\cite{avrahami2022blended}.
For \textit{single-view 3D reconstruction}, we adopt CLIP~\cite{radford2021learning} as the image encoder. 
For \textit{text-guided 3D generation}, we adopt BERT~\cite{devlin2018bert} as the text encoder.
The encoded features are then used to modulate the diffusion process with cross-attention.
\ifx \arxiv \undefined {
Please refer to the supplementary material for more details.
}
\fi

%
%




\begin{figure}[t!]
    \centering
     \includegraphics[width=\linewidth]{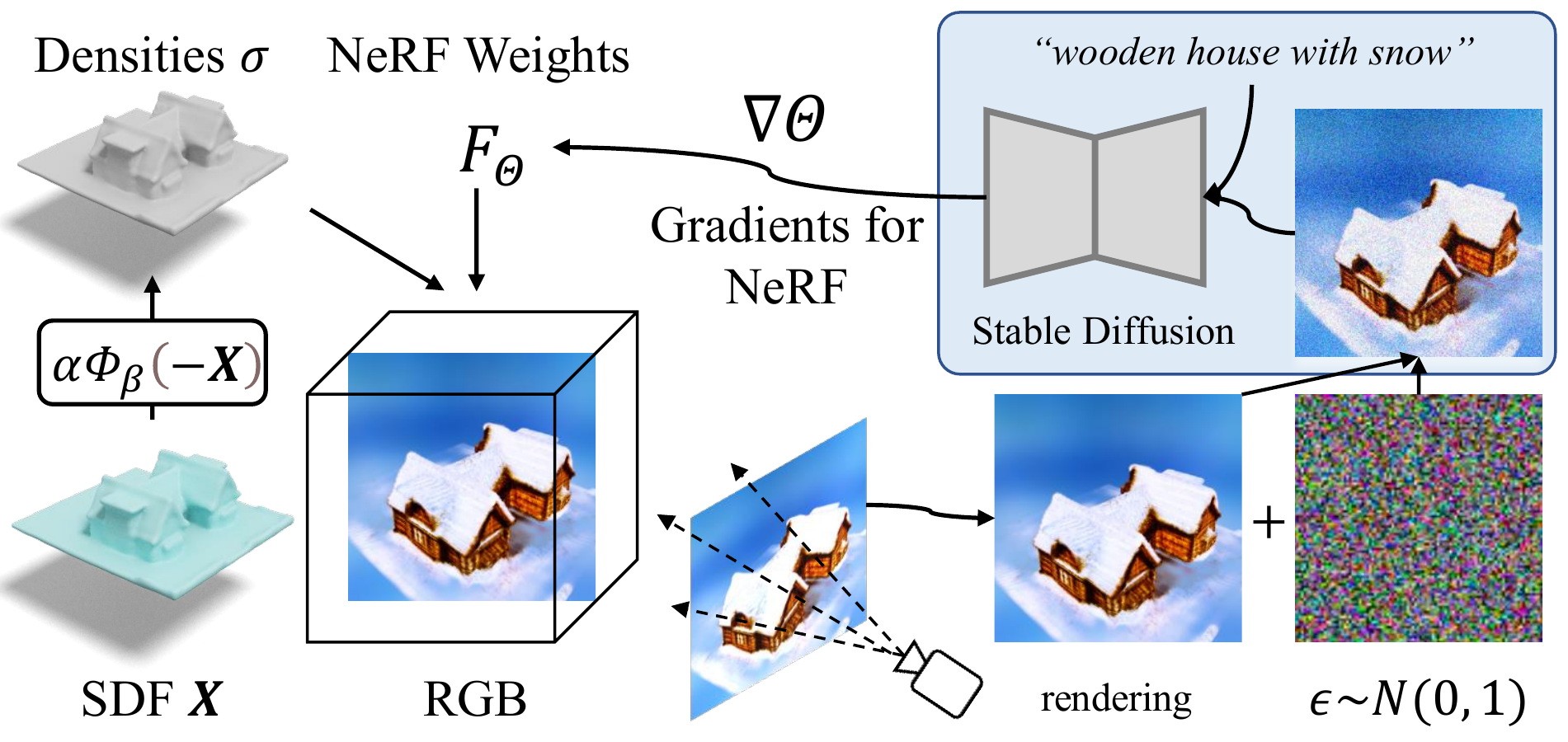}
    \vspace{-4mm}
    \caption{
    \textbf{3D Shape Texturing}.
    We demonstrate an application where models trained on 2D and 3D data are combined. 
    The shapes generated by \modelName are converted to a density tensor, then the color information is learned via neural rendering. The gradients are provided by an off-the-shelf 2D diffusion model~\cite{rombach2022high}.
    }

    \vspace \figmargin
    \vspace{-2mm}
    \figlabel{overview_p2}
\end{figure}

\begin{figure*}[t!]
    \centering
    \includegraphics[width=\linewidth]{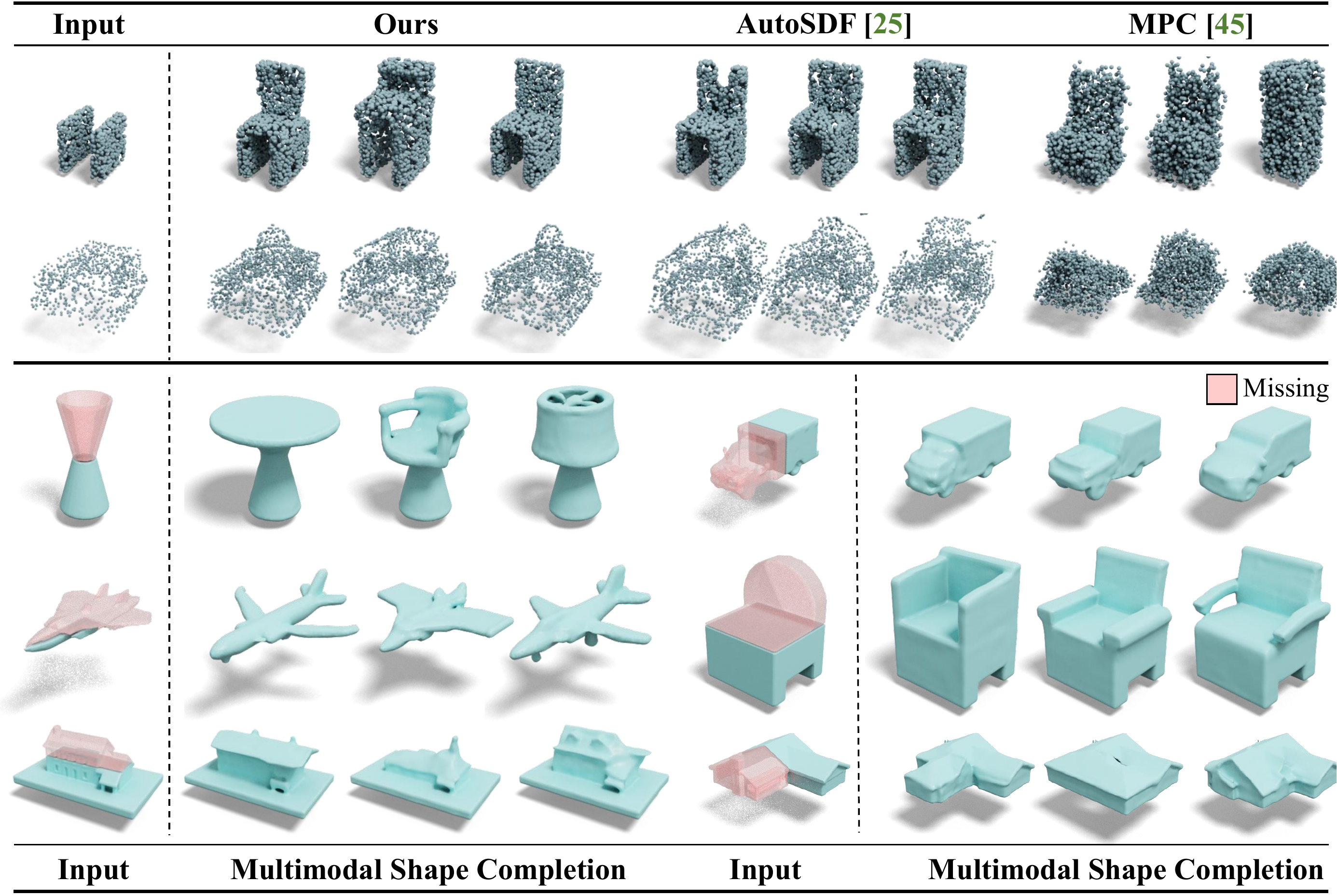}
    \vspace{-5mm}
    \caption{
    \textbf{Shape Completion.} \textit{(Top)} We compare \modelName with AutoSDF~\cite{autosdf2022} and MPC~\cite{wu2020multimodal} on  ShapeNet and BuildingNet data. \modelName generates shapes of better quality and diversity, while being consistent with the input partial shapes. We convert the generated SDFs from \modelName and AutoSDF to point clouds to compare with MPC. \ifx \arxiv \undefined { Please refer to the supplementary materials for SDF comparisons on BuildingNet for better visualization. } \fi
    \textit{(Bottom)} We present more results on diverse shape completion using various object categories.
    }
    \vspace{-3mm}
    \vspace \figmargin
    \figlabel{completion}
\end{figure*} 

\subsection{3D Shape Texturing with a 2D Model}
\label{subsec:texture}

While sampled 3D data  often exhibits compellingly detailed geometry,  textures of 3D data are generally more difficult to collect and even more often of limited quality.
Here, we  explore how to make the best use of 2D data and models, so as to aid 3D asset generation.
Thanks to the recent success of neural rendering~\cite{mildenhall2020nerf} and 2D text-to-image models trained on extremely large-scale data~\cite{rombach2022high}, using a 2D model to perform 3D synthesis is made possible with a score distillation sampling technique~\cite{poole2022dreamfusion}.

Here, we illustrate the procedure of texturing a 3D shape given a guiding input sentence $S$.
Starting from a generated T-SDF $\mathbf{X}$, we first convert it to a density tensor $\sigma$ by using VolSDF~\cite{yariv2021volume}.
Our goal is to learn a 5D radiance field to obtain color $F_\theta:(\mathbf{x},\mathbf{d})\rightarrow \mathbf{c}$, similar to the conventional NeRF~\cite{mildenhall2020nerf} setting. However, different from the conventional NeRF setting, the density tensor is fixed.
With a sampled camera pose, we can render an image $I$ by alpha-compositing the densities and colors along the rays for each pixel. 
Then, we distill the knowledge from a pre-trained stable diffusion model~\cite{rombach2022high}, denoted as a 2D UNet $\tilde{\epsilon}_\phi(\bfz_t,t,S)$ that predicts the noise at timestep $t$.
The most straightforward way to update $F_\theta$ is to obtain and backpropagate gradients all the way from the noise prediction loss term of the stable diffusion loss function to the input. However, the UNet Jacobian term is in practice expensive to compute. Therefore, it was proposed in~\cite{poole2022dreamfusion} to bypass the UNet and treat $\tilde{\epsilon}_\phi$ as a frozen critic providing scores by computing
\begin{equation}
    \mathbb{E}_{t,\epsilon}\left[\frac{\partial I(F_\theta(\cdot))}{\partial \theta}w(t) (\tilde{\epsilon}_\phi(\bfz_t,t,S)-\epsilon) \right],
\end{equation}
where $w$ is a time-dependent weight function defined in the stable diffusion model. 
The mechanism is called Score Distillation Sampling. Please refer to~\cite{poole2022dreamfusion} for details.
The score then provides an update direction to $F_\theta$.
We illustrate the process in \figref{overview_p2}.



\section{Experiments}
\seclabel{experiment}
In this section, we conduct extensive qualitative and quantitative experiments to demonstrate the efficacy and generalizability of \modelName. 
We evaluate methods on three tasks: shape completion, single-view 3D reconstruction, and text-guided generation. 
We then demonstrate two additional use cases: multi-conditional generation and 3D shape texturing. 

\subsection{Shape Completion}
\begin{table}[t]
\caption{
    \textbf{Quantitative comparison of Shape Completion}.
    We evaluate methods on fidelity (UHD) and diversity (TMD) using the ShapeNet and BuildingNet data.
    \modelName outperforms other methods in both metrics, especially diversity. 
}
\vspace{\tabmargin}
\vspace{-2mm}
\centering
\setlength{\tabcolsep}{0.25em}
\begin{tabular}{l cc cc} 
    \toprule
      &  \multicolumn{2}{c}{ShapeNet} & \multicolumn{2}{c}{BuildingNet}\\
    \cmidrule(r){2-3}\cmidrule(r){4-5}
    Method & \small{UHD $\shortdownarrow$} & \small{TMD $\shortuparrow$} & \small{UHD $\shortdownarrow$} & \small{TMD $\shortuparrow$}\\
    \midrule
    MPC~\cite{wu2020multimodal} & 0.0627 & 0.0303 & 0.1350 & 0.0467 \\
    AutoSDF~\cite{autosdf2022}  & 0.0567 & 0.0341 & 0.1208 & 0.0649 \\
    Ours  & \textbf{0.0557} & \textbf{0.0885} & \textbf{0.1116} & \textbf{0.0745} \\
    \bottomrule
\end{tabular}
\vspace{\tabmargin}
\label{tab:quan_shape_comp}
\end{table}

\begin{table}[t]
\caption{
    \textbf{Quantitative evaluation of single-view reconstruction}. We evaulate methods on the Pix3D dataset using Chamfer Distance and F-Score. We outperform other methods in both metrics.
}
\vspace{\tabmargin}
\vspace{-2mm}
\centering
\setlength{\tabcolsep}{0.25em}

\comment{
\begin{tabular}{l ccc} 
    \toprule
    {Method} & {IoU $\shortuparrow$ } & {CD $\shortdownarrow$ }  & {F-Score $\shortuparrow$ } \\
    \midrule
    Pix2Vox~\cite{xie2019pix2vox}      & 0.504 & 3.001 & 0.385 \\
    ResNet2TSDF     & 0.475 & 4.582 & 0.351  \\
    ResNet2Voxel   & 0.505 & 4.670 & 0.357 \\
    AutoSDF~\cite{autosdf2022}    & \textbf{0.521} & 2.267 & 0.415 \\
    Ours    & 0.487 & \textbf{1.185}& \textbf{0.432} \\
    \bottomrule
\end{tabular}
}
\begin{tabular}{l cc} 
    \toprule
    {Method} &  {CD $\shortdownarrow$ }  & {F-Score $\shortuparrow$ } \\
    \midrule
    Pix2Vox~\cite{xie2019pix2vox}    & 3.001 & 0.385 \\
    ResNet2TSDF  & 4.582 & 0.351  \\
    ResNet2Voxel    & 4.670 & 0.357 \\
    AutoSDF~\cite{autosdf2022}     & 2.267 & 0.415 \\
    Ours     & \textbf{1.852}& \textbf{0.432} \\
    \bottomrule
\end{tabular}
\vspace{\tabmargin}
\label{tab:quan_svr}
\end{table}

\begin{figure*}[t!]
\vspace{-0.5cm}
    \centering
     \includegraphics[width=\linewidth]{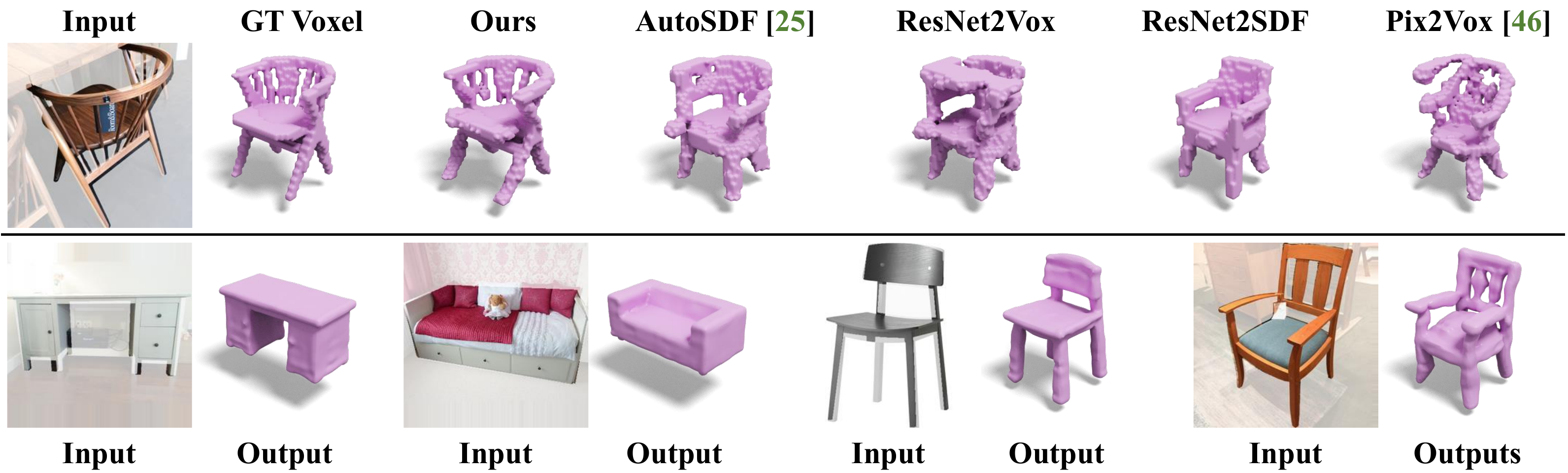}
    \vspace{-4mm}
    \caption{
    \textbf{Single-view 3D Reconstruction.} \textit{(Top)} We qualitatively compare all methods on the Pix3D dataset. \modelName generates shapes with the best visual quality. \textit{(Bottom)} Here we present more reconstruction results from \modelName.}
    \figlabel{image2shape_mm}
    \vspace \figmargin
\end{figure*} 

We evaluate the shape completion task on the ShapeNet~\cite{shapenet2015} and the BuildingNet~\cite{selvaraju2021buildingnet} datasets.
ShapeNet is a large-scale 3D CAD model dataset with 16 common object classes.
We use the train/test splits provided by Xu \etal~\cite{Xu2019disn}.
BuildingNet is a new large-scale 3D building model dataset. Compared to objects in ShapeNet, building models provide more geometric details and thus require higher resolution for representing the data. 
Hence, we use an SDF of $64^3$ resolution for ShapeNet, and a $128^3$ resolution for BuildingNet.
For both datasets, we compare the completion quantitatively by providing the bottom half of the ground truth shape as input, and evaluating the completed shapes generated by different methods.

\begin{figure*}[t!]
\vspace{-0.5cm}
    \centering
    \includegraphics[width=1.\linewidth]{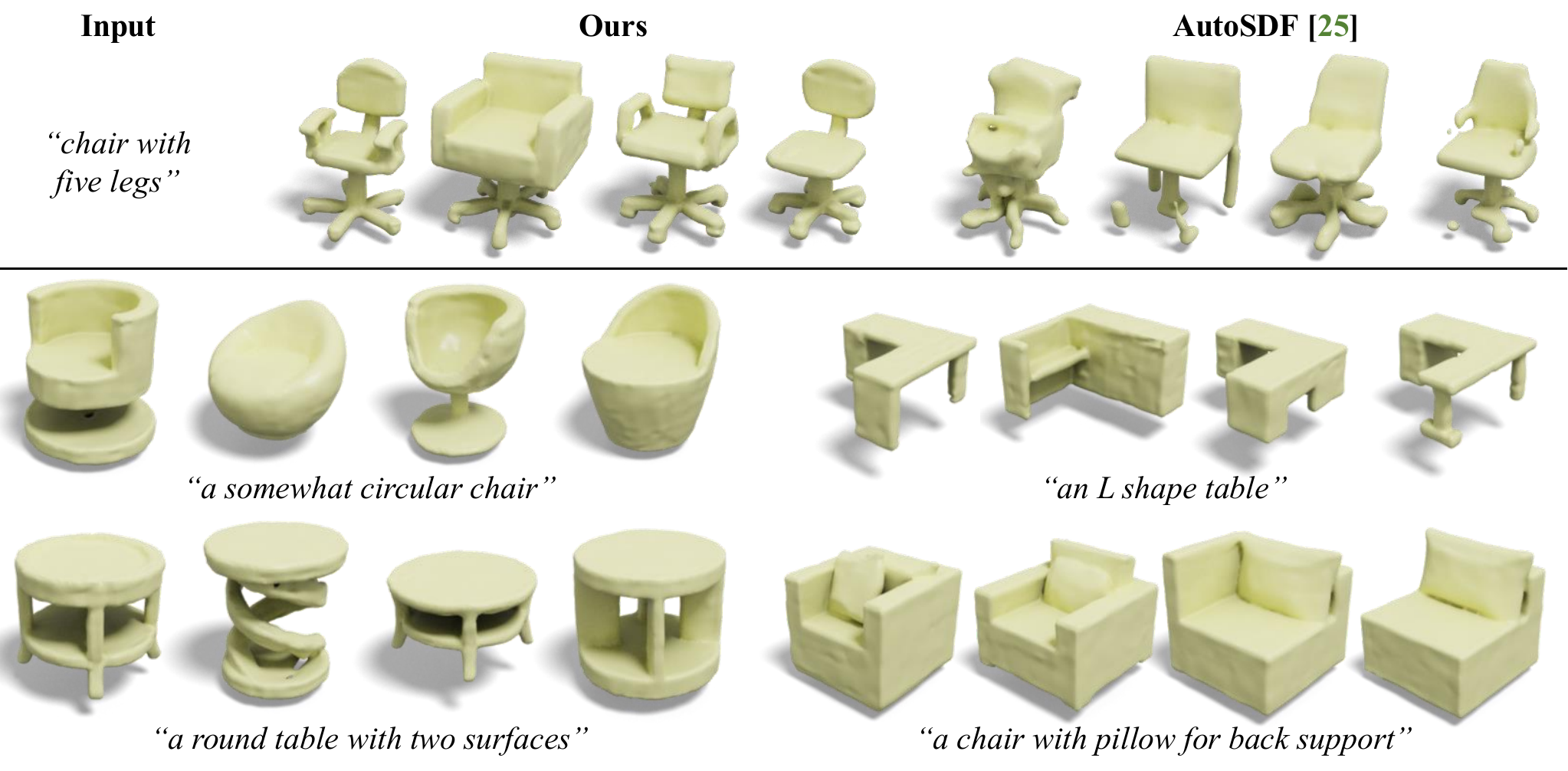}
     \vspace{-4mm}
    \caption{
    \textbf{Text-guided 3D shape generation.} \textit{(Top)} We compare \modelName with AutoSDF on the Text2Shape dataset. \modelName generates shapes with higher quality while conforming to the description. \textit{(Bottom)} We present more diverse and interesting text-guided generation results from \modelName.}
    \vspace \figmargin
    \vspace{-3mm}
    \figlabel{text2shape}
\end{figure*} 

We compare \modelName to the state-of-the-art point cloud completion method MPC~\cite{wu2020multimodal} and the autoregressive SDF generation method AutoSDF~\cite{autosdf2022}.
We adopt metrics from MPC~\cite{wu2020multimodal}. For each partial shape, we generate $k$ complete shapes. 
To evaluate \textit{completion fidelity}, we measure the Unidirectional Hausdorff Distance (UHD) between partial shapes and generated shapes.
To evaluate \textit{completion diversity}, we measure the Total Mutual Difference (TMD) by computing the average Chamfer distance among $k$ generated shapes. 
We use $k=10$ in the experiments. 

As shown in \tabref{quan_shape_comp}, the proposed \modelName performs favorably compared to all methods in the completion fidelity metric, and outperforms the baselines substantially in completion diversity.
The advantages of fidelity and diversity are also apparent in \figref{completion}.
Especially on the BuildingNet dataset, \modelName shows its advantages in modeling high-resolution and diverse data. AutoSDF and MPC struggle to model the distribution correctly.  

\begin{table}[t]
\caption{
    \textbf{Quantitative evaluation of text-guided generation}. We compare the proposed \modelName method, AutoSDF, and real data using a pretrained neural evaluator on the ShapeGlot dataset. $P$ denotes the neural evaluator preference rate for the target $P(\text{Tr})$ or the distractor $P(\text{Dis})$. If the preference is too close ($\leq 0.2$), we count the comparison as confused (conf.). Rows sum to $100\%$.
}
\vspace{\tabmargin}
\centering
\setlength{\tabcolsep}{0.25em}
\small
\begin{tabular}{cc cc cc} 
    \toprule
    Target~(Tr) & Distractor~(Dis) & $P(\text{Tr})$  & $P(\text{Dis})$ & $P(\text{conf.})$ \\
    \midrule
    Ours & AutoSDF~\cite{autosdf2022} & \textbf{49\%} & 36\% & 15\% \\
    Ours & GT & 33\% & \textbf{45\%} & 22\% \\
    AutoSDF~\cite{autosdf2022} & GT & 30\% & \textbf{49\%} & 21\% \\
    \bottomrule
\end{tabular}
\vspace{\tabmargin}
\vspace{-1mm}
\label{tab:quan_lang}
\end{table}

\begin{figure*}[t!]
\vspace{-0.0cm}
    \centering
    \includegraphics[width=\linewidth]{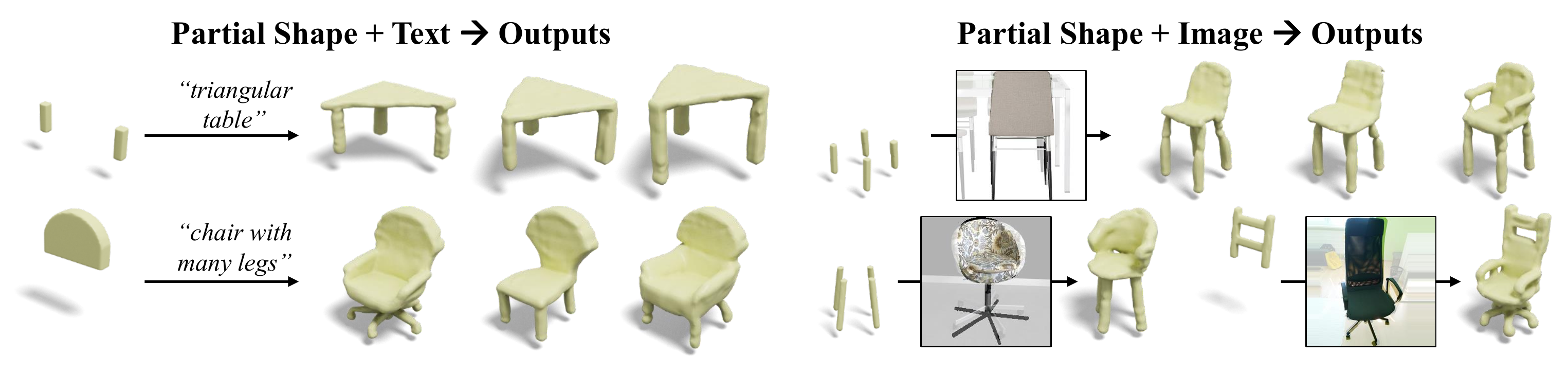}
    \vspace{-2mm}
    \caption{
    \textbf{Conditional generation from multiple modalities.}
    Here, we present generated samples from joint conditions of (left) partial shape and text and (right) partial shapes and images.
    The generated results are diverse and consistent with the provided conditions.
    }
    \vspace \figmargin
    \vspace{-3mm}
    \figlabel{multicond}
\end{figure*} 

\subsection{Single-view 3D Reconstruction}
\vspace{ \subsecmargin }

Next, we assess 3D shape reconstruction from a single image on the real-world benchmark Pix3D~\cite{sun2018pix3d} dataset.
We use the provided train/test splits on the chair category.
In the absence of official splits for other categories, we randomly split the dataset into disjoint train/test splits. 

We compare with the ResNet2TSDF and ResNet2Voxel baselines. Both encode images and directly output 3D shapes in the form of a T-SDF and a voxel grid. We also compare to two state-of-the-art methods for 3D reconstruction, \ie,  Pix2Vox~\cite{xie2019pix2vox} and AutoSDF~\cite{autosdf2022} .
We evaluate all methods after aligning resolutions to $32^3$ voxels. 
We use Chamfer Distance (CD), and F-score$@1\%$~\cite{what3d_cvpr19} as evaluation metrics. 

Quantitatively, as shown in \tabref{quan_svr}, \modelName outperforms other methods on both metrics.
Qualitatively, \modelName generates 3D shapes that are of higher visual quality and are more visually consistent with the objects shown in the images, regardless of camera poses, as shown in \figref{image2shape_mm}.

\begin{figure*}[t!]
\vspace{-0.5cm}
    \centering
     \includegraphics[width=.95\linewidth]{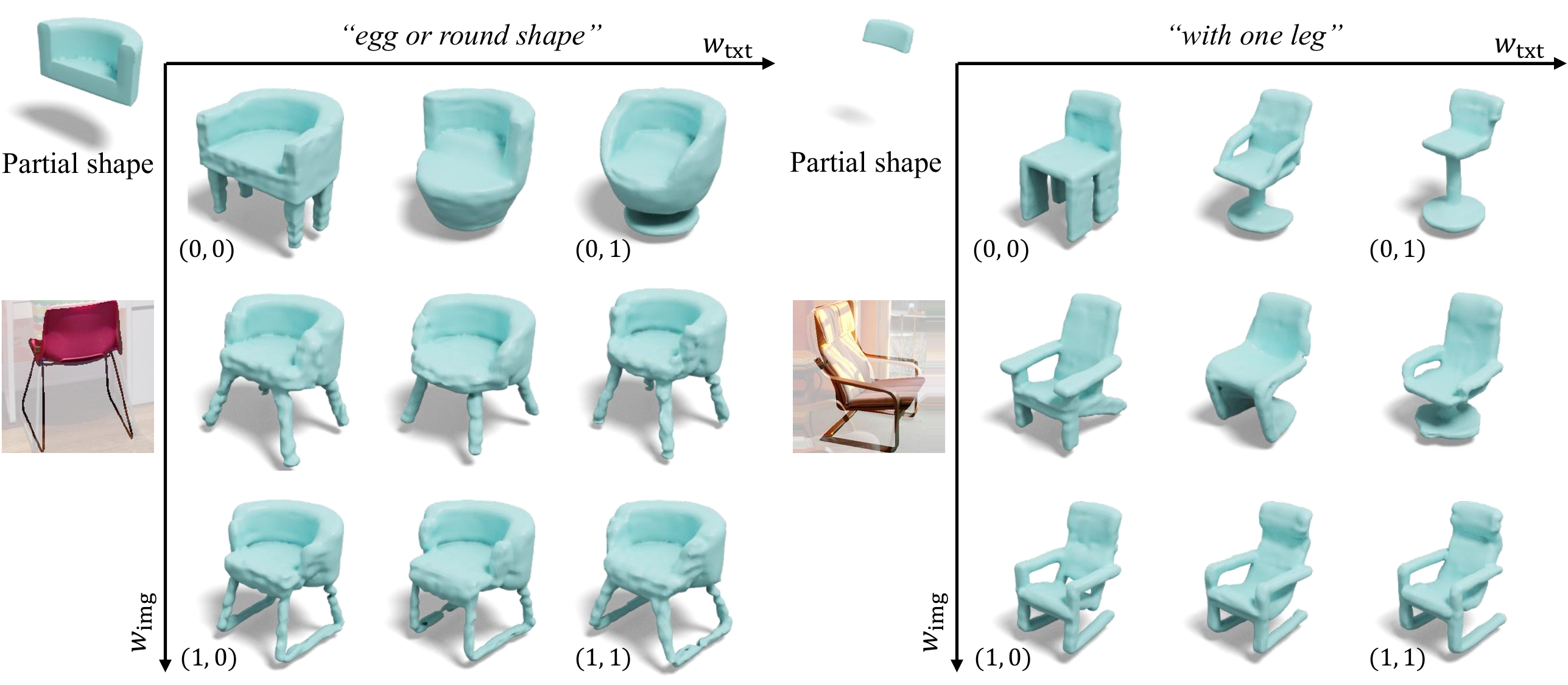}
    \caption{
    \textbf{Multiple conditioning variables with weight control.}
    Given a partial shape, an image, and a sentence as conditional input, we show that the model is sensitive to weights that control the importance of different modalities.
    }
    \vspace \figmargin
    \vspace{-2mm}
    \figlabel{multicond_poe}
\end{figure*} 
\begin{figure*}[t!]
    \centering
     \includegraphics[width=\linewidth]{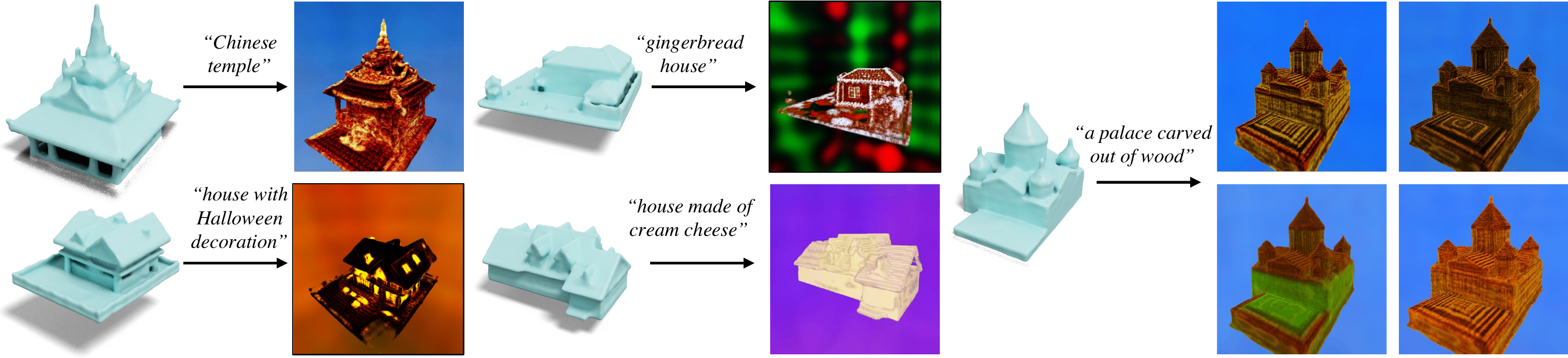}
    \caption{
    \textbf{3D Shape Texturing.}
    We texture the generated 3D shapes with a 2D diffusion model trained on large-scale data.
    This permits to generate textures from diverse textual inputs, including  style and material descriptions.
    The pipeline can also generate diverse results given the same input description.
    }
    \vspace \figmargin
    \vspace{-2mm}
    \figlabel{9_uncond_color}
\end{figure*} 

\subsection{Text-guided Generation}
\vspace{ \subsecmargin }

Next, we  evaluate  3D shape generation conditioned on text input. 
For a qualitative comparison, we use the Text2shape dataset~\cite{chen2018text2shape} that provides descriptions for the `chair' and `table' categories in  ShapeNet. For a quantitative evaluation, we adopt the ShapeGlot~\cite{achlioptas2019shapeglot} dataset which provides text utterances describing the difference between a target shape and two distractors based on the ShapeNet dataset. 
We compare \modelName with AutoSDF, which recently demonstrated state-of-the-art results on the text-guided 3D shape generation task. 

We follow the evaluation pipeline proposed by ShapeGlot~\cite{achlioptas2019shapeglot}.
We train a neural evaluator to distinguish the target shape from a distractor given the description. 
Given two shapes from different methods, the neural evaluator provides a confidence score for each of them based on the binary classification logits. 
For the absolute difference between two confidence scores $\le 0.2$, we count the comparison as confused (conf.).

As shown in \tabref{quan_lang}, \modelName quantitatively outperforms AutoSDF by a large margin with a low confusion rate. \modelName also performs better than AutoSDF when compared with ground truth data.
Qualitatively, we show in \figref{image2shape_mm} that \modelName not only generates shapes with better quality, but  that the generated shapes are also more diverse. 
Notably, \modelName reacts to very specific descriptions like ``L-shaped table'' and ``table with two surfaces.'' It generates objects of high diversity while remaining faithful to the provided description.

\subsection{Multi-conditional Generation}
\vspace{ \subsecmargin }

In addition to the conditional generation tasks which take a single conditioning variable as input, we further demonstrate the efficacy of \modelName in handling multiple modalities.
First, \modelName can jointly consider multiple conditioning modalities.
On the left of \figref{multicond}, we present the diverse generation conditional on both partial shapes and text. 
On the right of \figref{multicond}, we show that given partial shapes and images, \modelName can complete the different parts based on images. 
When there is ambiguity in images (\eg, rear-view of chairs), \modelName can produce diverse predictions.
Second, as shown in \figref{multicond_poe}, \modelName can not only be jointly conditioned on multiple inputs, but a weight can be used to control  the importance of the conditioning modalities, enabling more flexible user control.
For example, for the left sample in \figref{multicond_poe}, the larger the weight for the input image, the more similar the results are to the shapes in the image. Similarly, the larger the weight for input text, the more ``egg-shaped'' the results. We envision such a fine-grained form of control to be particularly useful for interactive user applications.

\subsection{3D Shape Texturing}
\vspace{ \subsecmargin }
Finally, we showcase an application that uses \modelName to generate 3D shapes of detailed geometry, and uses a pretrained text-to-image 2D diffusion model~\cite{rombach2022high} to provide textures. 
As shown in \figref{9_uncond_color}, the diffusion model pre-trained on large-scale 2D data can provide semantically meaningful and diverse guidance to texture the 3D shapes. 
The model shows superior expressiveness to interpret abstract concepts (\eg, Chinese- and Halloween-style) and materials (\eg, gingerbread, cream cheese). 
Given a single description, the texturing pipeline can also generate diverse results, as shown in the rightmost part of \figref{9_uncond_color}.

\section{Conclusion}
\vspace{ \secmargin }
\seclabel{conclusion}

In this work, we present \modelName, an attempt to adopt diffusion models to signed distance functions for 3D shape generation. 
To alleviate the computationally demanding nature of 3D representations, we first encode 3D shapes into an expressive low-dimensional latent space, which we use to train the diffusion model. 
To enable flexible conditional usage, we adopt class-specific encoders along with a cross-attention mechanism for handling conditions from multiple modalities, and leverage classifier-free guidance to facilitate weight control among modalities. 
Foreseeing the potential of a collaborative symbiosis between models trained on 2D and 3D data, we further demonstrate an application that takes advantage of a pretrained 2D text-to-image model to texture a generated 3D shape. 

Although the results look promising and exciting, there are quite a few future directions for improvement. First, \modelName is trained on high-quality signed distance function representations. To make the model more general and to enable the use of  more diverse data, a model that operates on various 3D representations simultaneously is desirable. Another future direction is related to the diversity of the data: we currently apply \modelName on object-centric data. It is interesting to apply the model to more challenging scenarios (\eg, entire 3D scenes). Finally, we believe there is  room to further explore how to combine models trained on 2D and 3D data. 

\noindent\textbf{Acknowledgements:}
 Work supported in part by NSF under Grants  2008387, 2045586, 2106825, MRI 1725729, and NIFA award 2020-67021-32799. Thanks to NVIDIA for providing a GPU for debugging.

\clearpage

{
\small
\bibliographystyle{ieee_fullname}
\bibliography{references}
}

\end{document}